\documentclass{article}

\usepackage{arxiv}

\usepackage[utf8]{inputenc} 
\usepackage[T1]{fontenc}    
\usepackage{hyperref}       
\usepackage{url}            
\usepackage{booktabs}       
\usepackage{amsfonts}       
\usepackage{nicefrac}       
\usepackage{microtype}      
\usepackage{lipsum}
\usepackage{graphicx}
\usepackage{subcaption}
\usepackage{xcolor}
\graphicspath{ {./images/} }
\usepackage{multirow}
\usepackage{placeins}
\newcommand{\modelname}{OpenUni}

\title{OpenUni: A Simple Baseline for Unified Multimodal Understanding and Generation}

\author{
\centerline{
Size Wu\thanks{Equal contribution.}~~\textsuperscript{\rm 1}\qquad
Zhonghua Wu\footnotemark[1]~~\textsuperscript{\rm 2}\qquad
Zerui Gong\footnotemark[1]~~\textsuperscript{\rm 1}\qquad
}\\
\centerline{
\textbf{
Qingyi Tao\textsuperscript{\rm 2} \qquad
Sheng Jin\textsuperscript{\rm 3} \qquad
Qinyue Li\textsuperscript{\rm 2} \qquad
Wei Li\textsuperscript{\rm 1}\qquad
Chen Change Loy\textsuperscript{\rm 1}
}}\\
\centerline{
\textsuperscript{\rm 1} S-Lab, Nanyang Technological University \quad
\textsuperscript{\rm 2} SenseTime Research \quad
\textsuperscript{\rm 3} SenseTime Research and Tetras.AI
}\\
\centerline{
\url{size001@e.ntu.edu.sg} \qquad 
\url{wei.l@ntu.edu.sg}  \qquad
\url{ccloy@ntu.edu.sg}
}
}

\begin{document}
\maketitle~\begin{abstract}
In this report, we present \emph{\modelname}, a simple, lightweight, and fully open-source baseline for unifying multimodal understanding and generation. Inspired by prevailing practices in unified model learning, we adopt an efficient training strategy that minimizes the training complexity and overhead by bridging the off-the-shelf multimodal large language models (LLMs) and diffusion models through a set of learnable queries and a light-weight transformer-based connector. With a minimalist choice of architecture, we demonstrate that \modelname~can: 1) generate high-quality and instruction-aligned images, and 2) achieve exceptional performance on standard benchmarks such as GenEval, DPG-Bench, and WISE, with only 1.1B and 3.1B activated parameters. To support open research and community advancement, we release all model weights, training code, and our curated training datasets (including 23M image-text pairs) at \url{https://github.com/wusize/OpenUni}.~\footnote[1]{This is an ongoing project.}
\end{abstract}

\begin{figure*}[h]
  \centering
  \begin{subfigure}[t]{0.49\textwidth}
    \centering
    \includegraphics[width=\linewidth]{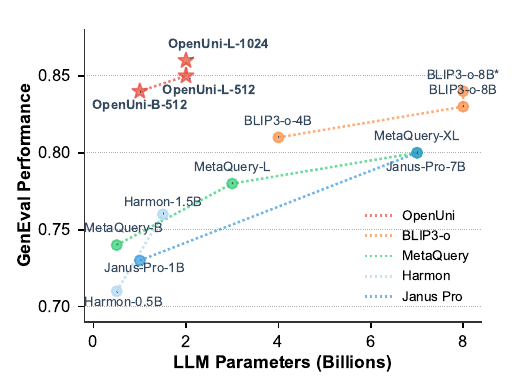}
    \caption{Evaluation of text-to-image generation: performance of \modelname\ variants and baselines on GenEval versus parameter count.}
    \label{fig:teaser-gen-eval}
  \end{subfigure}
  \hfill
  \begin{subfigure}[t]{0.49\textwidth}
    \centering
    \includegraphics[width=\linewidth]{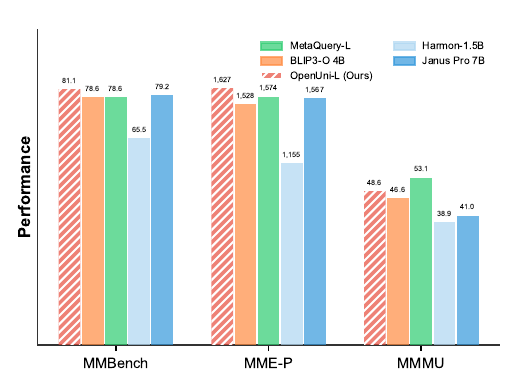}
    \caption{Multimodal understanding: comparison on \textsc{MMBench}, \textsc{MME-P} and \textsc{MMMU}.}
    \label{fig:teaser-understanding}
  \end{subfigure}

  \vspace{-4pt}  
  \caption{\modelname\ delivers strong performance with efficient parameter usage across both \emph{generation} and \emph{understanding} tasks.}
  \label{fig:teaser}
\end{figure*}

\section{Introduction}
The landscape of multimodal artificial intelligence has been dominated by the recent progress of multimodal large language models (LLMs)~\cite{zhu2023minigpt4enhancingvisionlanguageunderstanding, liu2023visualinstructiontuning, liu2024improved, li2024llava, instructblip, yang2024qwen2, chen2024internvl, chen2024far} and diffusion models~\cite{2022LDM, 2023Pixelartalpha, 2023SDXL, 2024lumina, 2024pixartsigma, ramesh2021zero, ramesh2022hierarchical, dalle3}, driven by architectural innovations and computational scalings of transformers~\cite{vaswani2017attention}. To further advance the frontier of multimodal intelligence, a natural leap forward would be integrating the two minds of understanding and generation into a single brain, demonstrated by GPT4-o~\cite{hurst2024gpt}'s impressive instruction-following ability of content generation.

Existing research efforts that unify multimodal understanding and generation can typically be divided into two subtracks. One type of work~\cite{team2024chameleon, wu2024vila, xie2024show, wu2024janus, chen2025janusprounifiedmultimodalunderstanding, li2024synergen, deng2025bagel} explores native multimodal models from scratch and shares the parameters of LLM for both tasks. Another line of work stitches LLMs and generation models to build unified frameworks~\cite{sun2024generative, wang2024illume, tong2024metamorph, chen2025blip3ofamilyfullyopen, wu2025harmonizing}. More recently, MetaQuery~\cite{pan2025transfermodalitiesmetaqueries} and BLIP3-o~\cite{chen2025blip3ofamilyfullyopen} directly align frozen multimodal LLMs with diffusion models, effectively instilling generation ability into an already established multimodal system. These works unveil the potential of a simple connection module that transfers the knowledge of well-trained LLMs to generation models, for controllable and high-quality visual generation.
Inspired by these studies, we present \emph{\modelname}, an open-source framework for unified multimodal understanding and generation, with minimum architectural complexities and computational overhead.

Specifically,
\modelname~adheres to the simplest design choices presented in MetaQuery~\cite{pan2025transfermodalitiesmetaqueries}, using only learnable queries and a light-weight connector between a multimodal LLM (MLLM) and a diffusion model. A two-stage training recipe is adopted to build \modelname. In the first (pre-training) stage, we align the LLM and diffusion model on 23M image-text pairs, by training only the learnable queries and the connector. The 23M training images used in this stage are sourced from public datasets and re-captioned by LLMs, which will also be released. In the finetuning stage, we unlock the diffusion model and train \modelname~on the 60k high-quality images contributed by BLIP3-o~\cite{chen2025blip3ofamilyfullyopen}.

We implemented three model variants, namely \modelname-B-512 and \modelname-L-512 and \modelname- L-1024, characterized by different model sizes and image resolutions. For image understanding, \modelname~inherits its base MLLM's strong performance on multimodal question-answering benchmarks. For image generation, our smaller variant \modelname-B-512 achieves a score of 0.84 on GenEval, on par with BLIP3-o-8B~\cite{li2023blip2bootstrappinglanguageimagepretraining}, using only 1.1B activated parameters, while significantly outperforming prior unified models like Janus-Pro~\cite{chen2025janusprounifiedmultimodalunderstanding}. Meanwhile, \modelname-L-1024 archives the best performance (0.86) among open-source unified models on GenEval with 3.1B activated parameters. Besides, \modelname~exhibits competitive performance on the WISE benchmark that assesses world knowledge comprehension, surpassing models that utilize LLMs of similar scales.
To support research and reproducibility, we release the full framework as an open-source baseline with minimal training complexity, modular design, publicly available training data, and an out-of-the-box training pipeline.

\section{Related Work}
\textbf{Multimodal LLMs for Visual Understanding.}
Built upon a visual encoder~\cite{clip, zhai2023sigmoid} and an LLM~\cite{touvron2023llama, touvron2023llama2, yang2024qwen2}, multimodal LLMs~\cite{zhu2023minigpt4enhancingvisionlanguageunderstanding, liu2023visualinstructiontuning,liu2024improved,li2024llava, zhu2025internvl3exploringadvancedtraining, wang2024qwen2, li2024mini, wu2024deepseek} produce language responses based on visual inputs, allowing new capabilities like visual reasoning, dialogue, and instruction following.
Despite their powerful visual understanding capabilities, most existing multimodal LLMs are limited to text outputs. Therefore, equipping these LLMs with visual generation ability would be the key step towards next-generation multimodal intelligence. In this work, we build \modelname~upon InternVL3~\cite{zhu2025internvl3exploringadvancedtraining}, adapting its original world knowledge to image generation.

\textbf{Text-to-Image Generation with Diffusion Models.}
Diffusion models~\cite{ramesh2021zero, 2022LDM, ramesh2022hierarchical, 2023SDXL, 2024pixartsigma, dalle3, peebles2023scalable, 2024hunyuandit} have become the dominant paradigm for image generation, producing high-quality visual content conditioned on language descriptions. Pioneering works~\cite{2022LDM, 2023SDXL} typically formulate image generation as Denoising Diffusion Probabilistic Models~\cite{ho2020denoising} (DDPM), based on a U-Net~\cite{ronneberger2015u} architecture. In more recent works~\cite{esser2024scalingrectifiedflowtransformers, xie2024sanaefficienthighresolutionimage, flux2024, ma2024sit}, Flow Matching (FM) forgoes explicit diffusion simulation by learning an ODE-driven continuous transformation from noise to data, effectively subsuming diffusion processes as a special case and permitting more direct probability transport paths to improve efficiency. For model architecture, state-of-the-art frameworks~\cite{peebles2023scalable, ma2024sit, zhuo2024lumina, flux2024, xie2024sanaefficienthighresolutionimage, xie2025sana} replace the U-Net backbone with diffusion transformers~\cite{vaswani2017attention} (DiTs). Among these models, SANA~\cite{xie2024sanaefficienthighresolutionimage, xie2025sana} enhances both training and sampling efficiency with increased compression ratio~\cite{chen2024deep} and linear attention~\cite{cai2023efficientvit}. In this work, we choose SANA as the diffusion module of \modelname~to reduce computation cost.

\textbf{Frozen LLMs.} Freezing the pre-trained weights of LLMs and incorporating task-specific modules has been an economic and effective approach to expanding their functionality~\cite{wu2024flmm, shi2024llamafusion, pan2025transfermodalitiesmetaqueries, chen2025blip3ofamilyfullyopen}. As an early attempt, F-LMM~\cite{wu2024flmm} builds a mask head on top of frozen LLMs, endowing the LLMs with grounding ability while preserving their reasoning and instruction-following capabilities. For image generation, LlamaFusion~\cite{shi2024llamafusion} introduces extra transformer modules for visual generation, alongside LLMs' original architecture. More recently, MetaQuery~\cite{pan2025transfermodalitiesmetaqueries} and BLIP3-o~\cite{chen2025blip3ofamilyfullyopen} effectively bridge multimodal LLMs and diffusion models with a set of learnable queries. Inspired by these works, \modelname~is built upon frozen multimodal LLMs. 

\textbf{Unifying Multimodal Understanding and Generation.}
There are two design philosophies regarding unifying multimodal understanding and generation. One subtrack~\cite{team2024chameleon, wu2024vila, xie2024show, wu2024janus, chen2025janusprounifiedmultimodalunderstanding, li2024synergen, deng2025bagel} explores native multimodal models from scratch and shares the parameters of LLM for both tasks. This type of work typically struggles to accommodate the two inherently heterogeneous tasks that require representations at different levels of granularity. Disentangled visual encoders~\cite{wu2024janus, chen2025janusprounifiedmultimodalunderstanding} or mixture of experts~\cite{li2024synergen, deng2025bagel, shukor2025scaling} are usually adopted to handle conflicting pathways.

The more resource-efficient approach stitches well-trained LLMs and generation models to build unified frameworks, connecting them with intermediate ViT features~\cite{sun2024generative, wang2024illume, tong2024metamorph, wu2025harmonizing} or learnable queries~\cite{pan2025transfermodalitiesmetaqueries, chen2025blip3ofamilyfullyopen}. Among these works, MetaQuery~\cite{pan2025transfermodalitiesmetaqueries} and BLIP3-o~\cite{chen2025blip3ofamilyfullyopen} build unified frameworks upon frozen multimodal LLMs, effectively transferring the knowledge learned in understanding tasks to visual generation. Our \modelname~follows the simple architecture introduced by MetaQuery~\cite{pan2025transfermodalitiesmetaqueries} and achieves significantly higher performance with fewer learnable parameters, setting a clean and strong baseline for this research direction.

\begin{table}[t]
    \centering
    \caption{Architecture specifications and number of training images of \modelname, MetaQuery~\cite{pan2025transfermodalitiesmetaqueries} and BLIP3-o~\cite{chen2025blip3ofamilyfullyopen}. *In BLIP3-o, the DiT that predicts CLIP features is regarded as the connector.}
    \scalebox{0.9}{
\begin{tabular}{l|c|c|c|c}
\hline
Model&MLLM  & \#Connector Params & Diffusion Model&\#Images \\ \hline
MetaQuery-B~\cite{pan2025transfermodalitiesmetaqueries} & LLaVA-OV-0.8B~\cite{li2024llava} & 316M & SANA-1.6B-512~\cite{xie2024sanaefficienthighresolutionimage}&25M \\ 
MetaQuery-L~\cite{pan2025transfermodalitiesmetaqueries} & Qwen2.5VL-3B~\cite{bai2025qwen2} & Unknown & SANA-1.6B-512~\cite{xie2024sanaefficienthighresolutionimage}&25M \\ 
MetaQuery-XL~\cite{pan2025transfermodalitiesmetaqueries} & Qwen2.5VL-7B~\cite{bai2025qwen2} & Unknown & SANA-1.6B-512~\cite{xie2024sanaefficienthighresolutionimage}&25M \\ 
BLIP3-o-4B~\cite{chen2025blip3ofamilyfullyopen} & Qwen2.5VL-3B~\cite{bai2025qwen2} & 1.4B* & SDXL (2.6B)~\cite{2023SDXL} & 30M \\ 
BLIP3-o-8B~\cite{chen2025blip3ofamilyfullyopen} & Qwen2.5VL-7B~\cite{bai2025qwen2} & 1.4B* & SDXL (2.6B)~\cite{2023SDXL} & 30M/50M \\ 
\hline
OpenUni-B-512 & InternVL3-1B~\cite{zhu2025internvl3exploringadvancedtraining} & 54M & SANA-0.6B-512~\cite{xie2024sanaefficienthighresolutionimage} & 23M\\ 
OpenUni-L-512 & InternVL3-2B~\cite{zhu2025internvl3exploringadvancedtraining} & 225M & SANA-1.6B-512~\cite{xie2024sanaefficienthighresolutionimage} & 23M \\ 
OpenUni-L-1024 & InternVL3-2B~\cite{zhu2025internvl3exploringadvancedtraining} & 225M & SANA-1.5-1.6B-1024~\cite{xie2025sana} & 23M \\ \hline
\end{tabular}

}
    \label{tab:model_specs}
\end{table}

\section{\modelname}
\subsection{Model}
Architecturally, \modelname~follows the design of MetaQuery~\cite{pan2025transfermodalitiesmetaqueries}, comprising $N$ learnable queries, a multimodal LLM, a transformer-based connector, and a diffusion model. In our implementation, we set $N=256$. The visual understanding ability of the multimodal LLM is fully retained since its weights remain frozen. During image generation, the learnable queries extract conditioning information from the user's prompt during the LLM's forward pass; this information is then processed by the connector and passed to the diffusion model via its cross-attention module. 

\textbf{Lightweight Connector.} The architecture of \modelname's connector is adapted from SigLIP's visual encoder~\cite{zhai2023sigmoid}. Different from pioneering works~\cite{pan2025transfermodalitiesmetaqueries, chen2025blip3ofamilyfullyopen} that feature a heavy connecting module between the LLM and diffusion model, \modelname's connector only comprises six transformer layers. 

\textbf{Model Variants.} In this work, we build three model variants. \modelname-B-512 is based on InternVL3-1B~\cite{zhu2025internvl3exploringadvancedtraining} and SANA-0.6B-512px~\cite{xie2024sanaefficienthighresolutionimage} while \modelname-L-512 adopts InternVL3-2B~\cite{zhu2025internvl3exploringadvancedtraining} and SANA-0.6B-512px~\cite{xie2024sanaefficienthighresolutionimage}. In addition, we increase the resolution of \modelname-L's image generation by changing the diffusion model to SANA-1.5-1.6B-1024px~\cite{xie2025sana}. This higher-resolution variant is named \modelname-L-1024. The model specifications of \modelname~are provided in Table~\ref{tab:model_specs}, alongside comparisons with the pioneering works MetaQuery~\cite{pan2025transfermodalitiesmetaqueries} and BLIP3-o~\cite{chen2025blip3ofamilyfullyopen}.

\textbf{Prompt Format.} For text-to-image generation, we use the following prompt template to format user instruction: ``\texttt{User: Generate an image <caption>\textbackslash n Assistant:}''. \texttt{<caption>} represents the image description. During training, \texttt{<caption>} is randomly set to empty for 10\% data samples to enable classifier-free guidance (CFG)~\cite{ho2022classifier} in inference.

\begin{table*}[t]
  \centering
\caption{Detailed hyperparameters in pre-training (stage I) and fine-tuning (stage II).}
\scalebox{0.9}
{\begin{tabular}{lcc}
\toprule
Setting & Stage I & Stage II \\
\midrule
Diffusion Model & Frozen & Trainable\\
Learning Rate & $ 10^{-4}$  & $10^{-5}$ \\
Batch Size & 512 & 256  \\
Optimizer & AdamW~\cite{loshchilov2017decoupled} & AdamW~\cite{loshchilov2017decoupled} \\

Grad. Clip &1.0 & 1.0\\
Weight Decay & 0.05& 0.05\\
Betas & (0.9, 0.95) & (0.9, 0.95) \\
Schedule & Cosine & Cosine \\
Training Steps & 100,000 & 10,000  \\
Warm-up Steps & 1,000 & 100 \\

\bottomrule
\end{tabular}}
\label{tab:hyperparameters}
\end{table*}

\subsection{Training Recipe}
We adopt a two-stage training strategy, where we first align the LLM and the diffusion model in a pre-training stage. Then we fine-tune the aligned modules using high-quality training data. The training hyperparameters are listed in Table~\ref{tab:hyperparameters}.

\begin{table}[t]
  \centering
\caption{Results from the GenEval benchmark for text-to-image generation. Here, BLIP3-o-8B* indicates the model that is trained with 30 million additional proprietary data samples. $\dagger$ marks results that are obtained by models trained on distillation data (i.e., BLIP3-o-60K~\cite{chen2025blip3ofamilyfullyopen}). $\ddagger$ denotes rewritten prompts. $\S$ stands for RL-based methods.}
\label{tab:geneval}
\resizebox{0.95\linewidth}{!}{
\begin{tabular}{llccccccc}
    \toprule
    \textbf{Type} & \textbf{Method} & \textbf{Single Obj.} & \textbf{Two Obj.} & \textbf{Counting} & \textbf{Colors} & \textbf{Position} & \textbf{Color Attri.} & \textbf{Overall$\uparrow$} \\
    \midrule
    \multirow{9}{*}{\textit{Gen. Only}} 
    & LlamaGen~\cite{sun2024autoregressive} & 0.71 & 0.34 & 0.21 & 0.58 & 0.07 & 0.04 & 0.32 \\
    & LDM~\cite{rombach2022highresolutionimagesynthesislatent} & 0.92 & 0.29 & 0.23 & 0.70 & 0.02 & 0.05 & 0.37 \\
    & SDv1.5~\cite{rombach2022highresolutionimagesynthesislatent} & 0.97 & 0.38 & 0.35 & 0.76 & 0.04 & 0.06 & 0.43 \\
    & PixArt-$\alpha$~\cite{chen2023pixart} & 0.98 & 0.50 & 0.44 & 0.80 & 0.08 & 0.07 & 0.48 \\
    & SDv2.1~\cite{rombach2022highresolutionimagesynthesislatent} & 0.98 & 0.51 & 0.44 & 0.85 & 0.07 & 0.17 & 0.50 \\
    & DALL-E 2~\cite{ramesh2022hierarchical} & 0.94 & 0.66 & 0.49 & 0.77 & 0.10 & 0.19 & 0.52 \\
    & Emu3-Gen ~\cite{wang2024emu3} & 0.98 & 0.71 & 0.34 & 0.81 & 0.17 & 0.21 & 0.54 \\
    & SDXL~\cite{2023SDXL}  & 0.98 & 0.74 & 0.39 & 0.85 & 0.15 & 0.23 & 0.55 \\
    & DALL-E 3~\cite{dalle3} & 0.96 & 0.87 & 0.47 & 0.83 & 0.43 & 0.45 & 0.67 \\
    & SD3-Medium~\cite{esser2024scalingrectifiedflowtransformers} & 0.99&0.94& 0.72 & 0.89 & 0.33 & 0.60 & 0.74 \\

   &FlowGRPO~\cite{liu2025flowgrpo}  &1.00 &0.99 &0.95 &0.92 &0.99 &0.86 & 0.95$\S$ \\
    
    \midrule
    \multirow{18}{*}{\textit{Unified}}
    & Chameleon~\cite{team2024chameleon} & - & - & - & - & - & - & 0.39 \\
    & SEED-X~\cite{ge2024seed} & 0.97 & 0.58 & 0.26 & 0.80 & 0.19 & 0.14 & 0.51 \\
    & LMFusion~\cite{shi2024llamafusion} & - & - & - & - & - & - & 0.63 \\
    & Show-o~\cite{xie2024show} & 0.95 & 0.52 & 0.49 & 0.82 & 0.11 & 0.28 & 0.68 \\
    & EMU3~\cite{wang2024emu3} & - & - & - & - & - & - & 0.66$\ddagger$ \\
    & TokenFlow-XL~\cite{liu2024world} & 0.95 & 0.60 & 0.41 & 0.81 & 0.16 & 0.24 & 0.63$\ddagger$ \\
    & Janus~\cite{wu2024janus} & 0.97 & 0.68 & 0.30 & 0.84 & 0.46 & 0.42 & 0.61 \\
    & Janus-Pro-1B~\cite{chen2025janusprounifiedmultimodalunderstanding} & 0.98 & 0.82 & 0.51 & 0.89 & 0.65 & 0.56 & 0.73 \\
    & Janus-Pro-7B~\cite{chen2025janusprounifiedmultimodalunderstanding}  & 0.99& 0.89 & 0.59 & 0.90 & 0.79 & 0.66 & 0.80 \\
    & Harmon-0.5B~\cite{wu2025harmonizing}  & 0.99& 0.80 & 0.57 & 0.87 & 0.55 & 0.48 & 0.71 \\
    & Harmon-1.5B~\cite{wu2025harmonizing} & 0.99& 0.86 & 0.66 & 0.85 & 0.74 & 0.48 & 0.76 \\
    &SimpleAR-0.5B-SFT~\cite{wang2025simplear} &  - & - & - & - & - & - & 0.53 \\
    &SimpleAR-0.5B-RL~\cite{wang2025simplear} &  - & - & - & - & - & - & 0.59$\S$ \\
    &SimpleAR-1.5B-SFT~\cite{wang2025simplear} &  - & - & - & - & - & - & 0.61 \\
    &SimpleAR-1.5B-RL~\cite{wang2025simplear} &  - & - & - & - & - & - & 0.63$\S$ \\
    & MetaQuery-B~\cite{pan2025transfermodalitiesmetaqueries} &  - & - & - & - & - & - & 0.74$\ddagger$  \\
    & MetaQuery-L~\cite{pan2025transfermodalitiesmetaqueries} &  - & - & - & - & - & - & 0.78$\ddagger$  \\
    & MetaQuery-XL~\cite{pan2025transfermodalitiesmetaqueries} &  - & - & - & - & - & - & 0.80$\ddagger$ \\
    & BLIP3-o-4B~\cite{chen2025blip3ofamilyfullyopen} & - & - & - & - & - & - & 0.81$\dagger$ \\
    & BLIP3-o-8B~\cite{chen2025blip3ofamilyfullyopen} & - & - & - & - & - & - & 0.83$\dagger$\\
    & BLIP3-o-8B*~\cite{chen2025blip3ofamilyfullyopen} & - & - & - & - & - & - & 0.84$\dagger$\\
    &BAGEL~\cite{deng2025bagel} & 0.99 &0.94 &0.81 &0.88 &0.64 &0.63 &0.82\\
    &BAGEL-Rewrite~\cite{deng2025bagel} &0.98 &0.95 &0.84 &0.95 &0.78 &0.77 &0.88 $\ddagger$ \\
    
  \cline{2-9}
    & \modelname-B-512 & 0.99 & 0.91 & 0.74 & 0.90 & 0.77 & 0.73 & 0.84$\dagger$ \\
    & \modelname-B-512 & 0.99 & 0.71 & 0.55 & 0.82 & 0.25 & 0.42 & 0.62 \\
    & \modelname-L-512 & 0.99 & 0.91 & 0.77 & 0.90 & 0.75 & 0.76 & 0.85$\dagger$ \\
    & \modelname-L-1024  & 0.99& 0.92 & 0.76 & 0.91 & 0.82 & 0.77 & 0.86$\dagger$ \\
    
    \bottomrule
\end{tabular}
}
\end{table}

\textbf{Stage 1: Pre-training.}
The primary goal of this stage is to train the learnable queries and the lightweight connector to effectively bridge the multimodal LLM and the diffusion transformer. Parameters of both the LLM and the diffusion model are frozen in this stage.
The connector learns to translate the LLM's output features (elicited by the 256 learnable queries) into conditioning signals that the diffusion model can interpret. We use a large composite dataset comprising several publicly available image/text collections: text-to-image-2M\cite{text2image2m_2024}, LAION-Aesthetic- 6M\cite{schuhmann2022laion}, Megalith-10M\cite{matsubara2024megalith10m}, RedCaps-5M\cite{desai2021redcaps}. All of these images are captioned by LLMs. This results in a pre-training corpus of roughly 23 million image-text pairs.

\textbf{Stage 2: High-Quality Finetuning.}
To refine the generative capabilities of the entire system (connector and diffusion model) for improved instruction adherence, image quality, and robustness to diverse prompts, we leverage the instruction tuning dataset released by BLIP3-o~\cite{chen2025blip3ofamilyfullyopen}.
The dataset consists of 60,000 high-quality image-text pairs generated by prompting GPT-4o with diverse captions and using models like DALL-E3 and Midjourney for image synthesis.

\section{Evaluation}

This section details the evaluation setup, benchmarks, and results for \modelname. We evaluated \modelname's capabilities in both image generation and multimodal understanding, comparing them against state-of-the-art models. Our evaluation aims to demonstrate \modelname's ability to achieve competitive performance with a simpler and light-weight architecture.

\begin{table}[t]
  \centering
  \caption{Results from the DPG-Bench for text-to-image generation. Here, BLIP3-o-8B* indicates that the model is trained with 30 million additional proprietary data samples. $\P$ denotes the model trained without BLIP3-o-60K~\cite{chen2025blip3ofamilyfullyopen}.}
  \label{tab:dpg_bench}
  \resizebox{0.86\linewidth}{!}{
    \begin{tabular}{llccccccc}
      \toprule
      \textbf{Type} & \textbf{Method} & \textbf{Global} & \textbf{Entity} & \textbf{Attribute} & \textbf{Relation} & \textbf{Other} & \textbf{Overall$\uparrow$} \\
      \midrule
      \multirow{10}{*}{\textit{Gen. Only}} 
      & SDv1.5~\cite{rombach2022highresolutionimagesynthesislatent}  & 74.63 & 74.23 & 75.39 & 73.49 & 67.81 & 63.18 \\
      & PixArt-$\alpha$~\cite{chen2023pixart} & 74.97 & 79.32 & 78.60 & 82.57 & 76.96 & 71.11 \\
      & Lumina-Next~\cite{2024lumina}  & 82.82 & 88.65 & 86.44 & 80.53 & 81.82 & 74.63 \\
      & SDXL~\cite{2023SDXL}  & 83.27 & 82.43 & 80.91 & 86.76 & 80.41 & 74.65 \\
      & Playground v2.5~\cite{2024PG2.5}  & 83.06 & 82.59 & 81.20 & 84.08 & 83.50 & 75.47 \\
      & Hunyuan-DiT~\cite{2024hunyuandit}  & 84.59 & 80.59 & 88.01 & 74.36 & 86.41 & 78.87 \\
      & PixArt-$\Sigma$~\cite{2024pixartsigma}  & 86.89 & 82.89 & 88.94 & 86.59 & 87.68 & 80.54 \\
      & Emu3-Gen~\cite{wang2024emu3}  & 85.21 & 86.68 & 86.84 & 90.22 & 83.15 & 80.60 \\
      & DALL-E 3~\cite{dalle3}  & \textbf{90.97} & 89.61 & 88.39 & \textbf{90.58} & 89.83 & 83.50 \\
      & SD3-Medium~\cite{esser2024scalingrectifiedflowtransformers}  & 87.90 & \textbf{91.01} & 88.83 & 80.70 & 88.68 & 84.08 \\
      \midrule
      \multirow{11}{*}{\textit{Unified}}
      
      & Show-o~\cite{xie2024show}  & - & - & - & - & - & 67.27 \\
      & Janus~\cite{wu2024janus}  & 82.33 & 87.38 & 87.70 & 85.46 & 86.41 & 79.68 \\
      & Janus-Pro-1B~\cite{chen2025janusprounifiedmultimodalunderstanding}  & 87.58 & 88.63 & 88.17 & 88.98 & 88.30 & 82.63 \\
      & Janus-Pro-7B~\cite{chen2025janusprounifiedmultimodalunderstanding}  & 86.90 & 88.90 & 89.40 & 89.32 & 89.48 & \textbf{84.19} \\
      
      & MetaQuery-B~\cite{pan2025transfermodalitiesmetaqueries}  & - & - & - & - & - & 80.04 \\
     & MetaQuery-L~\cite{pan2025transfermodalitiesmetaqueries}  & - & - & - & - & - & 81.10 \\
    & MetaQuery-XL~\cite{pan2025transfermodalitiesmetaqueries}  & - & - & - & - & - & 82.05 \\
      & BLIP3-o-4B~\cite{chen2025blip3ofamilyfullyopen}  & - & - & - & - & - & 79.36 \\
    & BLIP3-o-8B~\cite{chen2025blip3ofamilyfullyopen} & - & - & - & - & - & 80.73 \\
    & BLIP3-o-8B*~\cite{chen2025blip3ofamilyfullyopen} & - & - & - & - & - & 81.60 \\
      \cline{2-8}
    & \modelname-B-512  & 85.87 & 87.33 & 86.54 & 86.91 & 89.43 & 80.29 \\
    & \modelname-B-512 & 88.21 & 85.34 & 85.15 & 87.01 & 85.64 & 79.04$\P$  \\
      & \modelname-L-512 & 81.37 & 87.67 & 88.64 & 88.18 & \textbf{89.77} & 81.54 \\
    & \modelname-L-1024  & 87.01 & 90.02 & \textbf{89.63} & 90.28 & 88.62 & 83.08 \\
      \bottomrule
    \end{tabular}
  }
\end{table}
\begin{table}[t!]
  \centering
  \caption{Results from the WISE benchmark evaluating world knowledge in text-to-image generation. Here, BLIP3-o-8B* indicates the model that is trained with an additional 30 million proprietary data. We highlight the best results in \textbf{bold}.}
  \label{tab:wise}
  \resizebox{0.95\linewidth}{!}{
  \begin{tabular}{llccccccc}
    \toprule
    \textbf{Type} & \textbf{Method} & \textbf{Cultural} & \textbf{Time} & \textbf{Space} & \textbf{Biology} & \textbf{Physics} & \textbf{Chemistry} & \textbf{Overall$\uparrow$} \\
    \midrule
    \multirow{9}{*}{\textit{Gen. Only}} 
    & SDv1.5~\cite{rombach2022highresolutionimagesynthesislatent} & 0.34 & 0.35& 0.32&0.28 &0.29 &0.21 &  0.32\\
    & SDv2.1~\cite{rombach2022highresolutionimagesynthesislatent} & 0.30 & 0.38 &0.35 & 0.33 & 0.34&0.21 & 0.32 \\
    & Emu3-Gen ~\cite{wang2024emu3} & 0.34&0.45&0.48 &0.41 &0.45 &0.27 & 0.39 \\
    & FLUX.1-schnell~\cite{flux2024} & 0.39  &0.44  &0.50 & 0.31&0.44  &0.26  & 0.40 \\
    & SD3-Medium~\cite{esser2024scalingrectifiedflowtransformers} & 0.42  & 0.44 &0.48 &0.39  &0.47 &0.29 & 0.42 \\
    & SDXL~\cite{2023SDXL} & 0.43  & 0.48 &0.47  &0.44  &0.45 &0.27 & 0.43 \\
    &SD3.5-Large~\cite{esser2024scalingrectifiedflowtransformers} & 0.44 &0.50 &0.58  & 0.44&0.52 &0.31 & 0.46 \\
    & PixArt-$\alpha$~\cite{chen2023pixart} & 0.45  & 0.50& 0.48 & \textbf{0.49}&0.56 &0.34 &  0.47\\
    & FLUX.1-dev~\cite{flux2024} & 0.48  & \textbf{0.58} &0.62 &0.42  &0.51 & 0.35& 0.50 \\
    \midrule
    \multirow{15}{*}{\textit{Unified}}
    & Show-o~\cite{xie2024show} & 0.28 &0.40&0.48 &0.30&0.46 &0.30 & 0.35\\
    & Janus~\cite{wu2024janus} & 0.16 &0.26 &0.35 & 0.28 &0.30 & 0.14&  0.23\\
    & Janus-Pro-1.5B~\cite{chen2025janusprounifiedmultimodalunderstanding} & 0.20& 0.28&0.45 & 0.24 & 0.32& 0.16&  0.26\\
    & MetaQuery-B~\cite{pan2025transfermodalitiesmetaqueries} & 0.44 & 0.49 & 0.58 & 0.41 & 0.49 & 0.34 & 0.46 \\
    & MetaQuery-L~\cite{pan2025transfermodalitiesmetaqueries} & \textbf{0.56} & 0.57 & 0.62 & 0.48 & \textbf{0.63} & 0.42 & 0.55 \\
    & MetaQuery-XL~\cite{pan2025transfermodalitiesmetaqueries} & \textbf{0.56} & 0.55 & 0.62 & 0.49 & \textbf{0.63} & \textbf{0.41} & 0.55 \\
    & Harmon-1.5B~\cite{wu2025harmonizing} & 0.38 & 0.48 & 0.52 & 0.37 & 0.44 & 0.29 & 0.41 \\

    & BAGEL~\cite{deng2025bagel} &0.44 &0.55& 0.68& 0.44& 0.60 &0.39 &0.52\\
    
    & BLIP3-o-4B~\cite{chen2025blip3ofamilyfullyopen} & - & - & - & - & - & - & 0.50 \\
    & BLIP3-o-8B~\cite{chen2025blip3ofamilyfullyopen} & - & - & - & - & - & - & 0.52 \\
    & BLIP3-o-8B*~\cite{chen2025blip3ofamilyfullyopen} & - & - & - & - & - & - & \textbf{0.62} \\
          \cline{2-9}
    & \modelname-B-512 & 0.37 & 0.45 & 0.58 & 0.39 & 0.50 & 0.30 & 0.43 \\
    & \modelname-L-512 & 0.51 & 0.49 & 0.64 & 0.48 & \textbf{0.63} & 0.35 & 0.52 \\
    & \modelname-L-1024 & 0.49 & 0.53 & \textbf{0.69} & \textbf{0.49} & 0.56 & 0.39 & 0.52 \\
    \bottomrule
  \end{tabular}
  }
\end{table}

\subsection{Image Generation}
To assess the text-to-image generation capabilities of \modelname, we employ a range of established benchmarks focusing on prompt adherence, semantic alignment, and world knowledge. Specifically, \textbf{GenEval}~\cite{ghosh2024geneval} is employed to evaluate the model's proficiency in following complex textual prompts, focusing on generating images with correct object attributes, counts, positions, and colors. Results are reported across various categories such as single object, multiple objects, counting, colors, and position. We also use \textbf{DPG-Bench}~\cite{hu2024ella}, a benchmark designed to examine the intricate semantic alignment capabilities of text-to-image models using lengthy and dense prompts. For DPG-bench, we report scores across its defined categories (Global, Entity, Attribute, Relation, Other) and the overall score. Finally, \textbf{WISE}~\cite{niu2025wise} is employed to evaluate the model's incorporated world knowledge and reasoning capability within the context of image generation.

\begin{figure}[t]
    \centering
    \includegraphics[width=0.95\linewidth]{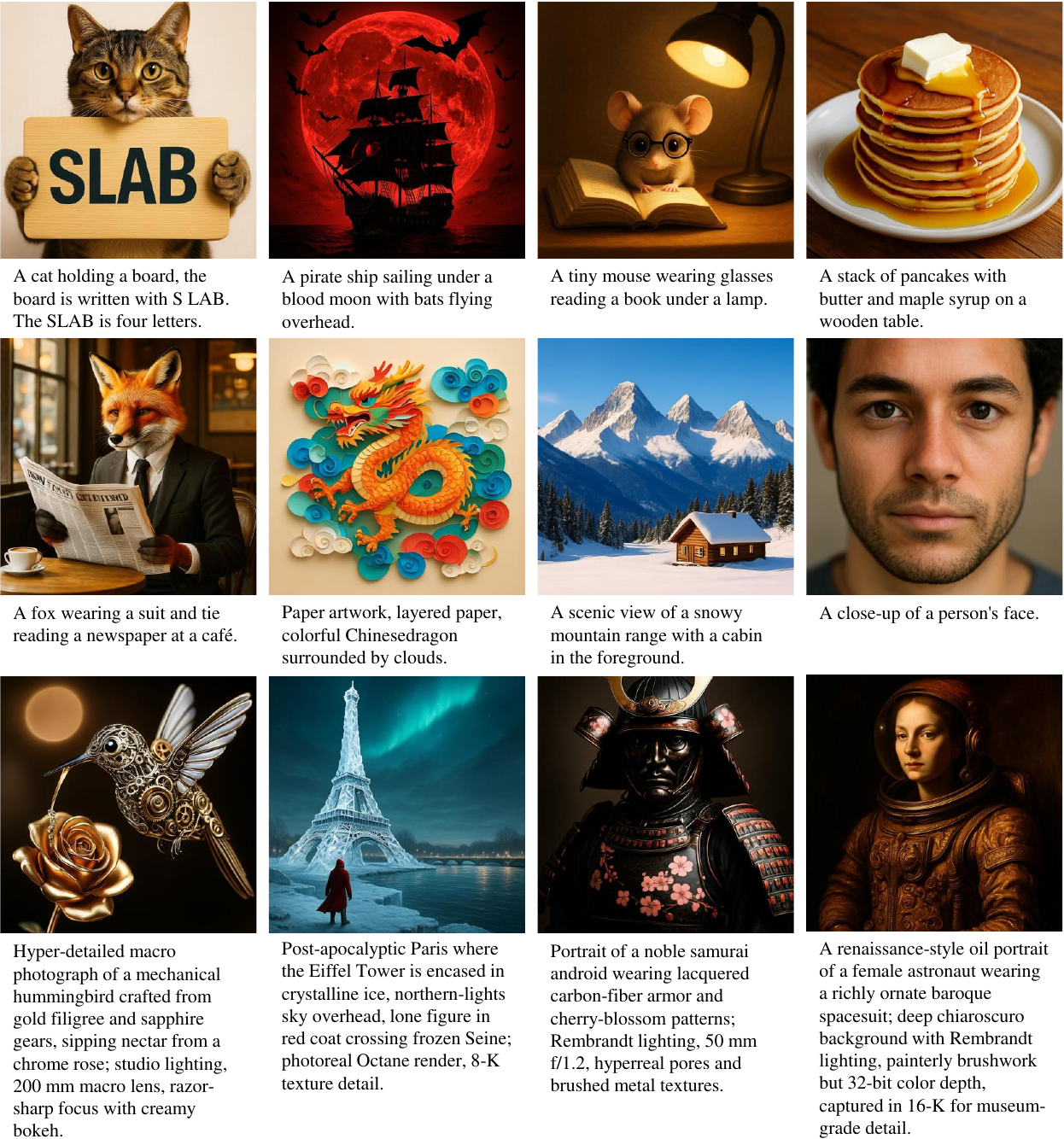}
    \caption{OpenUni is able to generate high-quality images and precisely follow user instructions. The above results are obtained by OpenUni-L-1024.}
    \label{fig:vis_gen}
\end{figure}

\begin{table*}[t!]
\centering
\caption{Results on image understanding benchmarks. Since the parameters of the MLLM (InternVL3~\cite{zhu2025internvl3exploringadvancedtraining}) are frozen, \modelname~preserves its excellent performance on the following benchmarks. We highlight the best results in \textbf{bold}.}
\fontsize{7pt}{8.8pt}\selectfont
\setlength\tabcolsep{6pt}
\renewcommand{\arraystretch}{1.1}
\resizebox{0.9\textwidth}{!}{
\begin{tabular}{l|cccccccc}
\toprule
Model & MMBench & SEED & MM-Vet & MME-P & MMMU & RWQA & TEXTVQA & POPE \\
\hline
EMU2 Chat \cite{sun2024generative} & - & 62.8 & 48.5 & - & 34.1 & - & 66.6 & - \\
Chameleon-7B \cite{team2024chameleon} & 19.8 & 27.2 & 8.3 & 202.7 & 22.4 & 39.0 & 0.0 & - \\
Chameleon-34B \cite{team2024chameleon} & 32.7 & - & 9.7 & 604.5 & 38.8 & 39.2 & 0.0 & - \\
Seed-X \cite{ge2024seed} & 70.1 & 66.5 & 43.0 & 1457.0 & 35.6 & - & - & - \\
VILA-U \cite{wu2024vila} & - & 59.0 & 33.5 & 1401.8 & - & 46.6 & 48.3 & 85.8 \\
LMFusion \cite{shi2024llamafusion} & 72.1 & 63.7 & - & 1603.7 & 41.7 & 60.0 & - & - \\
Show-o-512 \cite{xie2024show} & - & - & - & 1097.2 & 26.7 & - & - & 73.8 \\
EMU3~\cite{wang2024emu3} & 58.5 & 68.2 & 37.2 & - & 31.6 & 57.4 & 64.7 & 85.2 \\
MetaMorph \cite{tong2024metamorph} & 75.2 & 71.8 & - & - & - & 58.3 & 60.5 & - \\
TokenFlow-XL \cite{qu2024tokenflow} & 76.8 & 72.6 & 48.2 & 1551.1 & 43.2 & 56.6 & 77.6 & 86.8 \\
Janus-1.3B \cite{wu2024janus} & 69.4 & 63.7 & 34.3 & 1338.0 & 30.5 & - & - & 87.0 \\
Janus-Pro-7B~\cite{chen2025janusprounifiedmultimodalunderstanding} & 79.2 & 72.1 & 50.0 & 1567.1 & 41.0 & - & - & - \\
Harmon-0.5B~\cite{wu2025harmonizing} & 59.8 & 62.5 & - & 1148.0 & 34.2 & - & - & 86.5 \\
Harmon-1.5B~\cite{wu2025harmonizing} & 65.5 & 67.1 & - & 1155.0 & 38.9 & - & - & 87.6 \\
MetaQuery-B~\cite{pan2025transfermodalitiesmetaqueries} & 58.5 & 66.6 & 29.1 & 1238.0 & 31.4 & - & - & - \\
MetaQuery-L~\cite{pan2025transfermodalitiesmetaqueries} & 78.6 & 73.8 & 63.2 & 1574.3 & 53.1 & - & - & - \\
MetaQuery-XL~\cite{pan2025transfermodalitiesmetaqueries} & \textbf{83.5} & 76.9 & \textbf{66.6} & \textbf{1685.2} & \textbf{58.6} & - & - & - \\
BLIP3-O-4B~\cite{chen2025blip3ofamilyfullyopen} & 78.6 & 73.8 & 60.1 & 1527.7 & 46.6 & 60.4 & 78.0 & - \\
BLIP3-O-8B~\cite{chen2025blip3ofamilyfullyopen} & \textbf{83.5} & \textbf{77.5} & \textbf{66.6} & 1682.6 & 50.6 & \textbf{69.0} & \textbf{83.1} & - \\
\hline
\modelname-B (InternVL3-1B~\cite{zhu2025internvl3exploringadvancedtraining})& 72.6 & 58.2 & 59.5 & 1491.22  & 43.4 & 58.2 & 74.1 & \textbf{90.7} \\
\modelname-L (InternVL3-2B~\cite{zhu2025internvl3exploringadvancedtraining})& 81.1 & 64.6 & 62.2 & 1626.88 & 48.6 & 64.3 & 77.0 & 89.6 \\
\bottomrule
\end{tabular}
}
\label{tab:understanding_bench}
\end{table*}
The performance of \modelname~on the GenEval benchmark is presented in Table~\ref{tab:geneval}. It is remarkable that our smallest variant \modelname-B-512 (0.84) already matches the performance of larger models like MetaQuery-XL (0.80) and BLIP3-o-8B (0.84). \modelname-L-1024 archives an overall score of 0.86. Table~\ref{tab:dpg_bench} summarizes the performance of \modelname~on the DPG-Bench benchmark. On DPG-Bench, \modelname-L-1024 obtains an overall score of 83.08, surpassing all model variants of MetaQuery and BLIP3-o while being comparable to Janus-Pro-7B. For world knowledge evaluation on the WISE benchmark (Table ~\ref{tab:wise}), \modelname-L-512/1024 achieves 0.52, already matching the performance of BLIP3-o-8B trained on 30M public data. Finally, we visualize the image generation results in Figure~\ref{fig:vis_gen}.

\textbf{Discussion on GenEval.} Unlike general text-to-image benchmarks (e.g., DPG-Bench~\cite{hu2024ella}) that use free-form text prompts, GenEval~\cite{ghosh2024geneval} comes with a fixed prompt style: ``\texttt{a photo of A and B}''. The prior knowledge of the prompt format has made it much easier to distil GPT4-o~\cite{hurst2024gpt} or utilize RL methods~\cite{liu2025flowgrpo} on this benchmark. In our experiment on \modelname-B-512, we find utilizing the distillation data (i.e., BLIP3-o-60K~\cite{chen2025blip3ofamilyfullyopen}) drastically increases the result on GenEval from 0.62 to 0.84 as shown in Table~\ref{tab:geneval}. In contrast, we only observe a marginal performance gain on DPG-Bench (79.04 -> 80.29) as shown in Table~\ref{tab:dpg_bench}. This has raised concerns over data leakage in the community. Therefore, in Table~\ref{tab:geneval}, we labelled all results achieved by distillation data or RL methods, and report our \modelname's performance when the distillation data is not available. Additionally, we also mark the results obtained using rewritten prompts.

\subsection{Multimodal Understanding}
\begin{figure}[t]
    \centering
    \includegraphics[width=\linewidth]{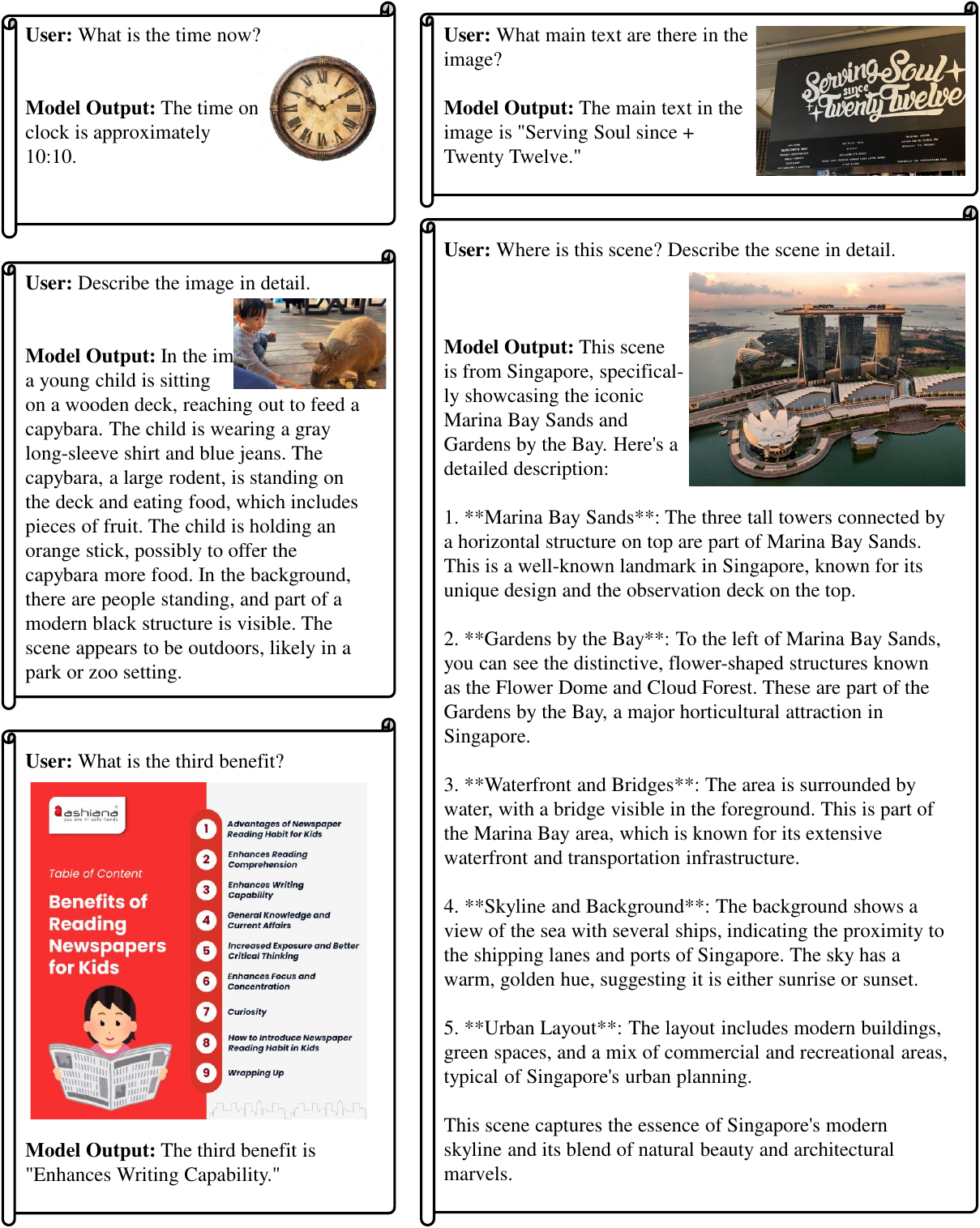}
    \caption{OpenUni has inherited InternVL3~\cite{zhu2025internvl3exploringadvancedtraining}'s excellent multimodal understanding ability to recognize visual patterns and comprehend general world knowledge. The results are obtained by OpenUni-L (based on InternVL3-2B).}
    \label{fig:vis_und}
\end{figure}

Since the frozen InternVL3~\cite{zhu2025internvl3exploringadvancedtraining} models are used to build \modelname, their core understanding capabilities are primarily inherited. We summarize their performance on several standard multimodal understanding benchmarks and compare with mainstream unified models. The reported benchmarks include MMBench, SEED-Bench, MM-Vet, MME-Perception (MME-P), MMMU, RealWorldQA (RWQA) and TextVQA. Here, we choose MMBench for evaluating diverse tasks requiring perception and reasoning; SEED-Bench for assessing generative comprehension; MM-Vet for evaluating integrated capabilities of large multimodal models; MME-Perception (MME-P) as a comprehensive benchmark for perception capabilities; MMMU for massive multi-discipline multimodal understanding and reasoning; RealWorldQA (RWQA) for assessing performance on real-world question answering; TextVQA which focuses on visual question answering where answers are present as text in the image; and POPE for evaluating object hallucination. As shown in Table~\ref{tab:understanding_bench}, \modelname~achieves competitive performance on these established benchmarks with only 1B and 2B activated parameters, thanks to InternVL3's outstanding visual perception and reasoning ability.
Finally, we show some examples of \modelname-L performing image understanding tasks in Figure~\ref{fig:vis_und}.

\section{Limitations}
In this work, we have introduced \modelname,~a simple but strong baseline for building unified models upon existing multimodal LLMs. As an ongoing project, \ \modelname~has the following limitations that will be addressed in future works: 1) the current \modelname~models struggle to render texts in generated images; 2) our largest model is based on a 2B LLM and 1.6B diffusion model. We believe scaling up the model sizes would further improve both understanding and generation performance of \modelname; 3) image-to-image generation tasks (e.g., reconstruction and editing) are left for future updates.

\FloatBarrier

\newpage
\bibliographystyle{unsrt}  
\bibliography{main}

\end{document}